\documentclass[sigconf, nonacm]{acmart}
\usepackage{multirow}
\newcommand\vldbyear{2024}
\newcommand\vldbworkshop{International Workshop on Quality in Databases (QDB'24)}
\newcommand\vldbauthors{\authors}
\newcommand\vldbtitle{\shorttitle} 
\newcommand\vldbavailabilityurl{}
\newcommand\vldbpagestyle{plain} 

\begin{document}
\title{AutoFAIR : Automatic Data FAIRification via Machine Reading}

\author{Tingyan Ma}
\affiliation{%
  \institution{Shanghai Jiao Tong University}
  \city{Shanghai}
  \country{China}
}
\email{xiaokeaiyan@sjtu.edu.cn}

\author{Wei Liu}
\affiliation{%
  \institution{Shanghai Jiao Tong University}
  \city{Shanghai}
  \country{China}
  \postcode{200240}
}
\email{liuw_515@sjtu.edu.cn}

\author{Bin Lu}
\affiliation{%
  \institution{Shanghai Jiao Tong University}
  \city{Shanghai}
  \country{China}
  \postcode{200240}
}
\email{robinlu1209@sjtu.edu.cn}

\author{Xiaoying Gan}
\authornote{Xiaoying Gan is the corresponding author.}
\affiliation{%
  \institution{Shanghai Jiao Tong University}
  \city{Shanghai}
  \country{China}
}
\email{ganxiaoying@sjtu.edu.cn}

\author{Yunqiang Zhu}
\affiliation{%
  \institution{Institute of Geographic Sciences and Natural Resources Research, Chinese Academy of Sciences}
  \country{China}
}
\email{zhuyq@igsnrr.ac.cn}

\author{Luoyi Fu}
\affiliation{%
  \institution{Shanghai Jiao Tong University}
  \city{Shanghai}
  \country{China}
  \postcode{200240}
}
\email{yiluofu@sjtu.edu.cn}
\author{Chenghu Zhou}
\affiliation{%
  \institution{Institute of Geographic Sciences and Natural Resources Research, Chinese
Academy of Sciences}
  \country{China}
}
\email{zhouch@igsnrr.ac.cn}

\begin{abstract}
The explosive growth of data fuels data-driven research, facilitating progress across diverse domains. The FAIR principles emerge as a guiding standard, aiming to enhance the findability, accessibility, interoperability, and reusability of data. However, current efforts primarily focus on manual data FAIRification, which can only handle targeted data and lack efficiency. To address this issue, we propose  AutoFAIR, an architecture designed to enhance data FAIRness automately. Firstly, We align each data and metadata operation with specific FAIR indicators to guide machine-executable actions. Then, We utilize Web Reader to automatically extract metadata based on language models, even in the absence of structured data webpage schemas. Subsequently, FAIR Alignment is employed to make metadata comply with FAIR principles by ontology guidance and semantic matching. Finally, by applying AutoFAIR to various data, especially in the field of mountain hazards, we observe significant improvements in findability, accessibility, interoperability, and reusability of data. The FAIRness scores before and after applying AutoFAIR indicate enhanced data value.
\end{abstract}

\maketitle

\pagestyle{\vldbpagestyle}
\begingroup\small\noindent\raggedright\textbf{VLDB Workshop Reference Format:}\\
\vldbauthors. \vldbtitle. VLDB \vldbyear\ Workshop: \vldbworkshop.\\ 
\endgroup
\begingroup
\renewcommand\thefootnote{}\footnote{\noindent
This work is licensed under the Creative Commons BY-NC-ND 4.0 International License. Visit \url{https://creativecommons.org/licenses/by-nc-nd/4.0/} to view a copy of this license. For any use beyond those covered by this license, obtain permission by emailing \href{mailto:info@vldb.org}{info@vldb.org}. Copyright is held by the owner/author(s). Publication rights licensed to the VLDB Endowment. \\
\raggedright Proceedings of the VLDB Endowment. 
ISSN 2150-8097. \\
}\addtocounter{footnote}{-1}\endgroup

\ifdefempty{\vldbavailabilityurl}{}{
\vspace{.3cm}
\begingroup\small\noindent\raggedright\textbf{VLDB Workshop Artifact Availability:}\\
The source code, data, and/or other artifacts have been made available at \url{\vldbavailabilityurl}.
\endgroup
}

\section{Introduction}

In the age of big data, the management and utilization of data have become pivotal for scientific advancement and industrial innovation. The FAIR principles \cite{wk2016FAIR}—Findability, Accessibility, Interoperability, and Reusability—have emerged as essential guidelines for enhancing the usability and value of data across diverse domains. These principles aim to address the challenges associated with inadequate metadata, and inefficient data discovery processes.

The introduction of the FAIR principles in 2016 \cite{wk2016FAIR} marked a significant shift in the acceptance and support of FAIR principles by various institutions. Subsequent efforts by the FAIR Metrics Working Group \cite{wk2018} led to a more comprehensive understanding of the FAIR principles, culminating in the proposal of evaluation metrics for FAIR compliance. From initial conceptual frameworks to practical implementations, the FAIR principles have gained wide popularity in research communities and institutions around the world. Initiatives like GO FAIR and FAIRsFAIR have played crucial roles in promoting standardized approaches to data management, emphasizing the importance of FAIR compliance for accelerating scientific discovery and innovation.

For domain-specific data, such as agriculture, healthcare, and climate, data publishers have developed specialized data FAIRification strategies tailored to each field. In the healthcare, the strategies \cite{health} emphasize the privacy and security of sensitive patient data while promoting interoperability between different health information systems, facilitating better clinical decision-making and research outcomes. These specialized strategies highlight the importance of detailed FAIRification methods to effectively implement the FAIR principles and enhance data utility across various scientific disciplines.

However, despite the advancements in FAIR adoption, practical challenges remain in technology to automate the handling process.  
\textbf{1. Complex Standards :} Despite the detailed implementation steps proposed by various groups, achieving uniformity across diverse data sources remains a significant challenge. \textbf{2. Lack of Automate Technology :}  Automating the FAIRification process poses technological challenges. Current methods often require manual intervention, especially for domain-specific data. These manual processes lack efficiency and scalability, hindering the broader adoption and consistent application of FAIR principles. 

To address the above issues, we propose an automated FAIRification architecture AutoFAIR, which converts non-FAIR compliant data into a FAIR-compliant format. By integrating Web Reader and FAIR Alignment, AutoFAIR can automatically process data, reducing the need for manual intervention while improving data FAIRness. The architecture not only complies with the FAIR Principles, but can also be executed automately to realize efficient processing and management of data from multiple sources. The application of automated processes not only improves the efficiency of data processing, but also greatly expands the scope of application of the FAIR principle and promotes the progress of  data sharing and utilization. Our main contributions are as follows:

\begin{itemize}
    \item We introduces AutoFAIR, an innovative architecture that automates the process of making dataset webpages FAIR-compliant.
    \item We propose a novel two-stage information extraction method in Web Reader which automatically extracts metadata from diverse data sources using web structure analysis and language models (LMs) to ensure comprehensive metadata generation even without standardized schemas.
    \item We design FAIR Alignment that employs ontology guidance and semantic matching to standardize information, converting heterogeneous metadata into a unified FAIR data profile.
    \item A case study in mountain hazard research demonstrate the effectiveness of AutoFAIR. By applying AutoFAIR, the FAIR scores of different websites have been significantly improved. Based on the FAIR data profile, it generates spatiotemporal distribution maps that enhance data reusability and support keyword searches, providing data support for research.
\end{itemize}

\section{Related Work}
The FAIR Principles were first proposed by M.D. Wilkinson et al. in 2016 \cite{wk2016FAIR} and have been widely promoted and applied in subsequent years. The FAIR Indicators Working Group\cite{wk2018} conducted a comprehensive study and eventually proposed evaluation indicators for FAIR compliance. A.Jacobsen et al. further interpreted the FAIR Principles \cite{jaco}, highlighting the key points to consider during implementation.  Additionally, M. Hahnel, V. Dan, A. Dunning, et al. evaluated selected data repositories \cite{10, 40}, analyzing their performance concerning data FAIRness. To more accurately assess the extent to which data repositories have implemented the FAIR principles, organizations such as the FAIR Indicators Group \cite{wk2018}, the Dutch Data Archiving and Networking Service (DANS) \cite{thomas_fair}, the Australian Research Data Sharing Organization (ANDS) \cite{ands_fair}, FAIRsFAIR \cite{fsf} and Open Science Cloud Europe \cite{rd_fair} have proposed their own assessment models and tools. These tools typically contain clear assessment indicators and scoring mechanisms, providing an objective basis for the management and improvement of data repositories.

Some methods have been proposed on how to implement the FAIR principles. A. Jacobsen et al. proposed a generic process for make digital resources FAIR \cite{5}, while GO FAIR \cite{goFAIR} proposes a full set of methods and tools for the FAIRification of data in a particular domain [6]. In addition, several groups and organizations have published recommendations to guide the implementation of FAIR, such as the European Commission's "Making FAIR a reality" report, which provides 27 detailed recommendations and actions for different stakeholders \cite{7}. The European Open Science Cloud FAIR Working Group has also made six recommendations to advance the implementation of FAIR \cite{8}.  These generic processes cover all aspects of data creation, storage, sharing, and reuse, emphasizing the importance of standardization and normalization of operations. However, despite these detailed implementation steps, practical challenges remain in technology to automate the handling process. For domain-specific data, such as agriculture \cite{argri}, healthcare \cite{health}, and climate \cite{climate}, data publishers have developed specialized data FAIRification strategies based on their respective domains' characteristics. However, these manual methods can only perform on targeted data and lack cross-domain and automation support.

Web information extraction has long been a critical area of study, focusing on efficiently extracting information from webpages. Traditional approaches predominantly rely on the HTML text of webpages, leveraging the DOM tree's graph structure or serialized processing methods. Approaches such as SimpDOM\cite{simpdom} and FreeDOM\cite{freedom} exploit the inherent tree structure of the DOM for node classification, while Transformer-based methods like DOM-LM\cite{domlm} and MarkupLM\cite{markuplm} integrate text and markup within a unified framework. 
 In contrast, the Web Reader component of AutoFAIR introduces a novel two-stage information extraction method that simultaneously considers the DOM tree structure and textual semantics. This ensures comprehensive metadata extraction without the need for structured data schemas, facilitating further compliance with FAIR principles through the FAIR Alignment module. 

 \begin{table}[h]
  \caption{At the data level, we extract data information required by the FAIR principles; at the metadata level, we complete the metadata profile according to the FAIR principles. Each operation corresponds to one or more FAIR indicators.}
  \label{tab:rule}
  \renewcommand\arraystretch{1.3}
  \begin{tabular}{ccl}
    \toprule
    Level & Operation & Indicators\\
    \midrule
     \multirow{5}*{data} & \parbox[t]{5cm}{1. extract the permanent unique identifier of the data} & \parbox[t]{1.3cm}{\centering F1}\\
~ & \parbox[t]{5cm}{2. extract the core elements of the data (creator, title, publisher, release date, abstract, keywords, etc.)} & \parbox[t]{1.3cm}{\centering F2, R1.2}\\
~ & \parbox[t]{5cm}{3. extract the license} & \parbox[t]{1.3cm}{\centering R1.1}\\
\midrule
\multirow{8}*{metadata} & \parbox[t]{5cm}{1. metadata contains identifiers for the data it describes} & \parbox[t]{1.3cm}{\centering F3}\\
~ & \parbox[t]{5cm}{2. metadata is stored in the data platform to make it machine-retrievable } & \parbox[t]{1.3cm}{\centering F4, A2, A1}\\
~ & \parbox[t]{5cm}{3. metadata contains retrieval information about the data and is periodically checked to see if the data is accessible} & \parbox[t]{1.3cm}{\centering A1}\\
~ & \parbox[t]{5cm}{4. metadata is standardized using the DCAT model} & \parbox[t]{1.3cm}{\centering I1, R1.3}\\
~ & \parbox[t]{5cm}{5. metadata contains links between the data and other entities} & \parbox[t]{1.3cm}{\centering I3}\\
  \bottomrule
\end{tabular}
\end{table}

\section{Preliminary}

As an initial effort to automate data FAIRification, we establish a domain-agnostic data handling process to facilitate the discovery and reuse of data. We observe that in the implementation of FAIR principles, the most important guidelines evolve into the scoring details of FAIRsFAIR \cite{fsf} through step-by-step development \cite{wk2016FAIR, wk2018, goFAIR, FAIR_data_maturity}. We improve the FAIRness of data based on the FAIRsFAIR metrics, and in \autoref{tab:rule} we show the operations at the data and metadata level and also illustrate the correspondence of each operation to the FAIR metrics. 

At the data level, we focus on extracting permanent unique identifiers, core elements such as creator, title, publisher, release date, and keywords, as well as license information. These operations correspond to specific FAIR metrics, ensuring that each aspect of the data is properly accounted for and documented. At the metadata level, we aim to complete the metadata profile by embedding identifiers, ensuring machine retrievability, and standardizing descriptions using models like DCAT. Additionally, we emphasize the periodic verification of metadata accessibility and the inclusion of links to other entities, which are crucial for maintaining data interoperability and reusability.
\begin{figure*}
  \centering
  \includegraphics[width=\linewidth]{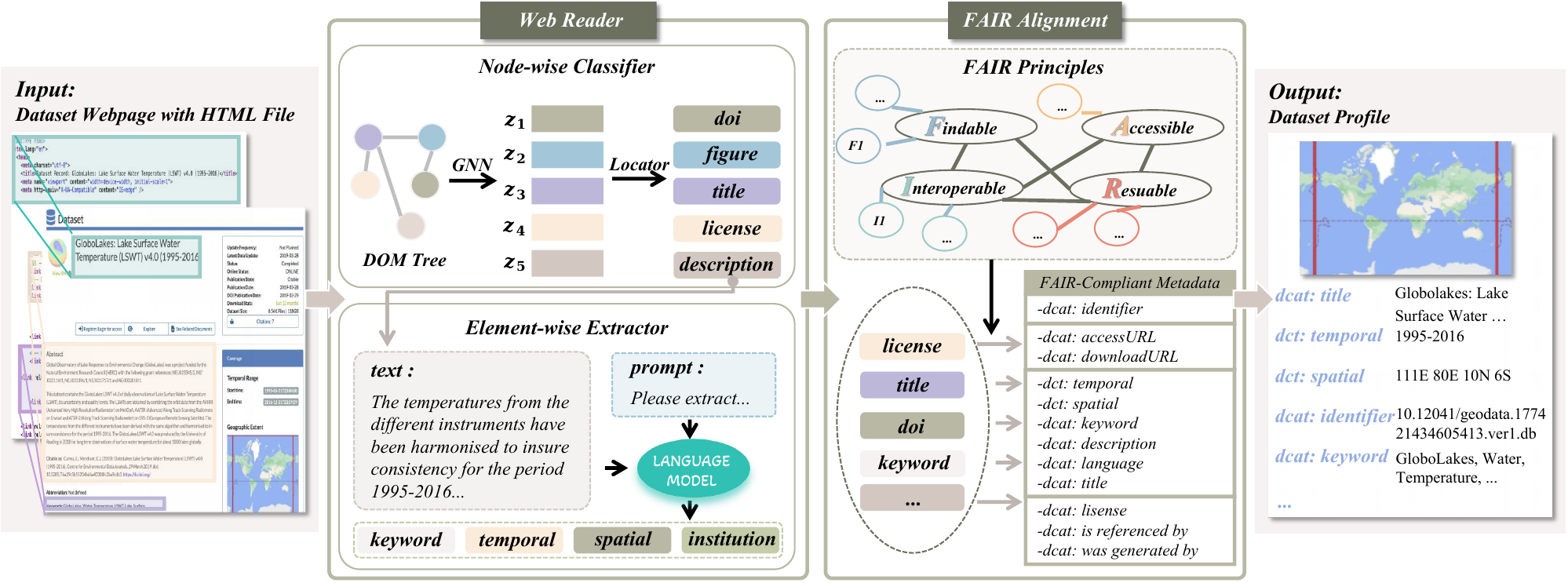}
  \caption{Overview of AutoFAIR’s Architecture. The DOM tree is constructed from the data webpage HTML. In Web Reader, nodes are categorized by a graph neural network to locate metadata fields, and for nodes with long text, a language model extracts the metadata. The extracted fields are then mapped according to the FAIR principles through FAIR Alignment, resulting in a FAIR-compliant metadata profile.}
  \label{fig:fs}
\end{figure*}

\section{Method}
\subsection{Overall Architecture}
AutoFAIR extracts and standardizes metadata from data webpages using two main components: Web Reader and Fair Alignment module, showed in \autoref{fig:fs}. Web Reader converts the webpage HTML into a DOM tree and uses Graph Neural Networks (GNNs) to classify nodes, locating the desired metadata. For nodes with complex text, a language model extracts the most relevant information. Fair Alignment module standardizes fields through ontology guidance and semantic matching, creating data entries on the DataExpo website \cite{dataexpo} and embedding DCAT \cite{dcat} metadata to ensure machine-readability and unified findability across sites. Additionally, the system processes temporal and spatial information using techniques like coordinate transformation, geographic registration, and interpolation to support spatiotemporal map search.

\subsection{Web Reader}
Firstly, we convert the HTML of the data webpage into a DOM tree, where each node corresponds to an element in the HTML. Next, a node-wise classifier utilizing graph neural networks is employed to locate the desired metadata information. For nodes containing complex text, we use a language model to extract the most relevant information. Through this pipeline, we can effectively obtain the metadata of the data.

\subsubsection{Node-wise Classifier}

The process of converting an HTML document into a DOM tree involves several steps:
\begin{enumerate}
    \item \textbf{Tokenization}: The HTML document is tokenized into a sequence of tokens representing HTML elements, attributes, and text.
    \item \textbf{Tree Construction}: A DOM tree is constructed recursively from the HTML tokens. Each HTML element becomes a node in the tree, with its attributes stored as properties of the node. Child elements become children of their parent nodes in the tree.
\end{enumerate}

Once the HTML document is represented as a DOM tree, we apply Graph Neural Networks (GNNs) model to classify nodes within the tree. Let $\mathbf{X}$ be the node feature matrix of size $|V| \times d$, where $|V|$ is the number of nodes and $d$ is the dimension of the node features. Each row of $\mathbf{X}$ corresponds to the feature vector of a node in the DOM tree. At each layer of the GNN, the message passed from node $v_j$ to node $v_i$ is computed as:

\begin{equation}
\mathbf{m}_{ij}^{(l)} = \text{ReLU}(\mathbf{W}^{(l)} \mathbf{h}_j^{(l-1)}),
\end{equation}
where $\mathbf{h}_j^{(l-1)}$ is the representation of node $v_j$ at the $(l-1)$-th layer, $\mathbf{W}^{(l)}$ is a learnable weight matrix. After computing the messages, they are aggregated to update the node representations:

\begin{equation}
\mathbf{h}_i^{(l)} = \text{AGGREGATE} \left( \left\{ \mathbf{m}_{ij}^{(l)} \mid v_j \in \mathcal{N}(v_i) \right\} \right),
\end{equation}
where $\mathcal{N}(v_i)$ represents the set of neighboring nodes of $v_i$. Finally, the node representations are fed into a softmax classifier to predict the label probabilities:

\begin{equation}
p(y_i \mid v_i) = \text{softmax}(\mathbf{W}^{(K)} \mathbf{h}_i^{(K)}),
\end{equation}
where $K$ is the number of GNN layers and $\mathbf{W}^{(K)}$ is the weight matrix of the classifier.

The parameters of the classifier are learned by minimizing the cross-entropy loss over the labeled nodes:

\begin{equation}
\mathcal{L} = -\sum_{(v_i, y_i) \in L} \sum_{c=1}^{|Y|} \mathbb{I}(y_i = c) \log p(y_i = c \mid v_i),
\end{equation}
where $\mathbb{I}(\cdot)$ is the indicator function.

\subsubsection{Element-wise Extractor}
Once the node-wise classifier identifies the HTML nodes corresponding to the metadata fields, we utilize language models to extract the relevant information from these nodes. Specifically, we take the text content of each identified HTML node and use the BERT model to encode this text, guiding the extraction process with a carefully designed prompt. For instance, when we need to extract key spatiotemporal information from a dataset description, the prompt is tailored to emphasize the importance of these details. The final piece of relevant information is then obtained by processing the output from BERT.

\subsection{Fair Alignment}
To address the challenges posed by inconsistent information descriptions within FAIR principles and their varied formats, we adopt techniques such as ontology guidance and semantic matching. These methods aim to standardize each field. To establish a comprehensive data index and enhance data interoperability, we align each field by creating a dataset entry on the DataExpo website \cite{dataexpo} and embedding DCAT \cite{dcat} metadata within the page. This strategic approach ensures the provision of machine-readable metadata, enabling unified dataset searches across multiple sites.

\[
DCAT\_Metadata = f(\text{Original\_Metadata}).
\]

For temporal and spatial alignment, which is crucial for effective spatiotemporal map retrieval, we employ techniques like coordinate transformation, geographic registration, and interpolation. Let \( T_i \) denote temporal information in diverse formats and \( S_i \) represent spatial information in various coordinate systems. We define functions \( f_t \) for temporal standardization, \( f_s \) for spatial transformation, \( f_g \) for geographic registration, and \( f_i \) for interpolation. This ensures alignment across all fields, with standardized temporal and spatial data represented as \( t_u \) and \( s_u'' \) respectively:

\[
t_u = f_t(T_i), \quad s_u'' = f_i(f_g(f_s(S_i))).
\]
\subsection{Theoretical Analysis of FAIR Principles' Implementation in AutoFAIR}
AutoFAIR's design is rooted in the FAIR principles (Findability, Accessibility, Interoperability, and Reusability). The Web Reader component enhances Findability by extracting comprehensive metadata from webpages using Graph Neural Network (GNNs) and language models that analyzes the DOM structure for accurate metadata extraction. This metadata is made Accessible by structuring it in a machine-readable format. The FAIR Alignment component ensures Interoperability through ontology guidance and semantic matching, aligning metadata with widely accepted standards such as DCAT. This alignment guarantees that metadata can be easily integrated and used across different systems. Lastly, the comprehensive and standardized metadata enhances Reusability, making it easier for researchers to discover, access, and reuse data. 

\begin{figure*}
  \centering
  \includegraphics[width=1.0\linewidth]{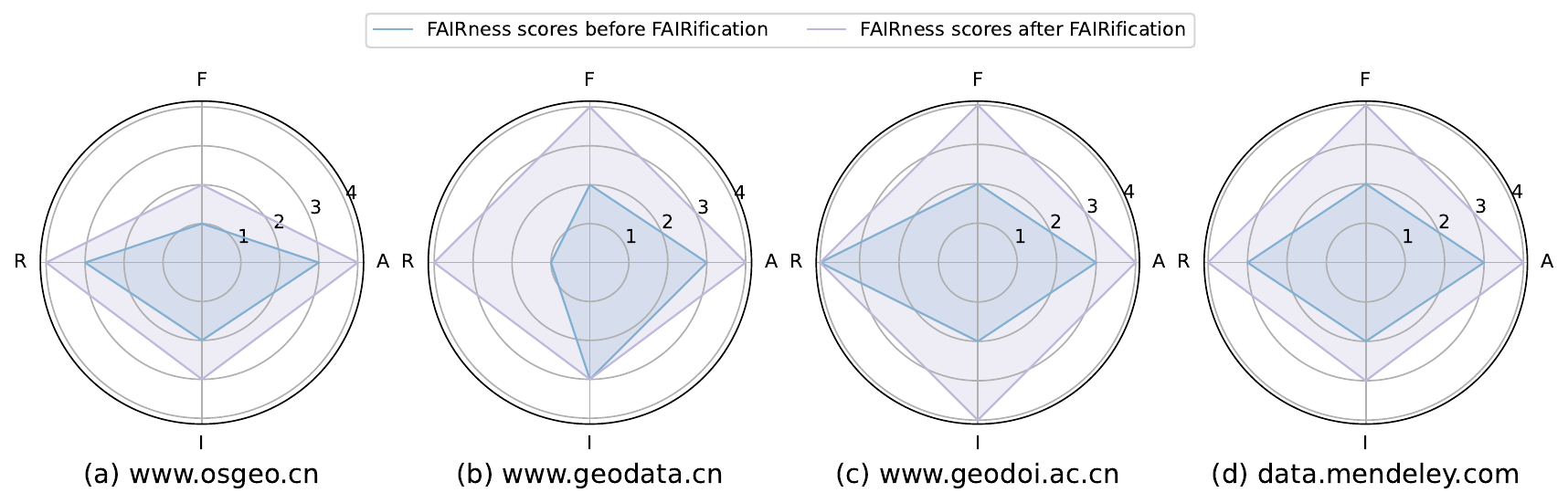}
  \caption{FAIRness scores for data under four domains before and after AutoFAIR. (a) and (b) conform to type 2 metadata type, i.e., metadata nested in html structure, and FAIRness is mainly enhanced by node-wise classifier extraction; (c) and (d) conform to type 3 metadata type, and element-wise extraction is required in addition to node-wise categorization to enhance FAIRness.}
  \label{fig:FAIRification}
\end{figure*}

\section{Data FAIRness Analysis}

\subsection{Datasets}
To analyze the effectiveness of our architecture, we take the field of mountain hazard as an example and collect data from various websites within this domain for data FAIRification. Specifically, we analyze 7124 data from 512 domains. The metadata in mountain hazard domains exhibits three key characteristics:
\begin{enumerate}
    \item \textbf{Metadata adheres to FAIR principles:} Data webpages align with FAIR principles, ensuring findability, accessibility, interoperability, and reusability of associated resources.
    \item \textbf{Metadata embeds in HTML structures:} Data webpages with metadata embedded within HTML structures, increasing the difficulty of machine retrieval of data.
    \item \textbf{Metadata scatters in textual descriptions:} Data webpages with metadata fragments dispersed within textual descriptions, requiring sophisticated extraction techniques.
\end{enumerate}

In response to the situation where the metadata of the last two types does not meet the FAIR principles, we have successfully implemented AutoFAIR to extract webpage information, thereby enhancing the machine-readable metadata profile.

\subsection{The Impact of FAIRification on Dataset Fairness}
Based on the FAIRsFAIR \cite{fsf} released in 2020, we conducted an assessment of adherence to FAIR principles, comparing the FAIRness scores of the original data domains with those processed through AutoFAIR. Different from manual domain-oriented processing methods, the automated workflow of AutoFAIR enables it to efficiently process data webpages adaptively, covering a wide range of domains, and improving the FAIRness of data. However, the improvements achieved through our FAIRification process are limited and depend on the data information presented on the original webpages. 

\begin{figure}[h]
  \centering
  \includegraphics[width=0.8\linewidth]{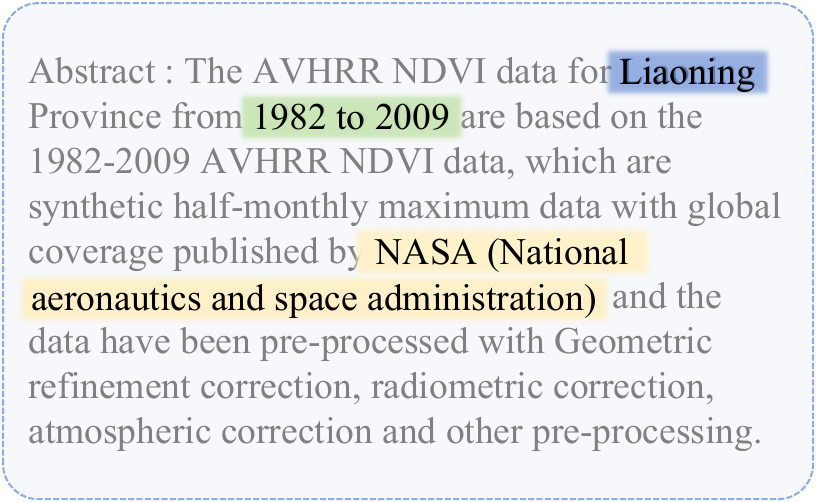}
  \caption{For this descriptive text embedded in the data webpage, spatiotemporal information and institutions can be fully extracted by Web Reader.}
  \label{fig:embed}
\end{figure}

\autoref{fig:FAIRification} illustrates the improvement in FAIRness scores on sampled domains. The amount of data contained under each domain ranges from 40 to 2216. In general, since these domains are well-established, their metadata information is comprehensive; however, their presentation methods do not meet the requirements of FAIR principles, resulting that reuse and machine utilization require targeted analysis of website structure. The metadata profiles obtained through our AutoFAIR process circumvent this issue of targeted handling. Through web reader, we fully extract metadata information nested within the structure and textual descriptions. Subsequently, through FAIR alignment, metadata documents are provided with standardized descriptions and formats. With our approach, accessibility and reusability are significantly enhanced. However, for some domains, the improvement in findability and interoperability limited due to the absence of the permanent unique identifier and license.

\autoref{fig:embed} shows a data webpage under "geodoi.ac.cn", which contains rich metadata information, but due to the fact that this information is nested in the text, such as metadata information in the abstract, it is not easy to be found and understood by machine when it is used. In our AutoFAIR process, the spatiotemporal information as well as the institutions are effectively extracted and then integrated into the metadata document to enhance the use value of data.

\subsection{Findable and Accessible}
Based on the powerful information extraction capability of Web Reader, we standardize the extraction of fields into metadata profiles that adhere to the FAIR principles. This effort enhances the findability and accessibility of data, as the FAIR metadata profiles contain information such as data titles, descriptions, keywords, and in some cases, references to other datasets. This improvement is showcased through the data website built on AutoFAIR.

\begin{table}[h]
  \caption{The number of datasets retrieved based on the Keyword "collapse" is more than regular databases.}
  \label{tab:acc}
  \begin{tabular}{ccl}
    \toprule
    Website & \#Dataset\\
    \midrule
     AutoFAIR & 58\\
     Findata & 28\\
     Google Dataset Search & 13\\
     Dimensions & 26\\
  \bottomrule
\end{tabular}
\end{table}

When searching for mountain hazard data using "collapse" as a keyword, our website has more data related to "collapse" compared to mainstream and commonly used data retrieval services. This finding partly reflects the improvement in the findable and accessible scores.

Further analysis reveals that this improvement is primarily attributed to our comprehensive extraction of webpage information and FAIR alignment. Web Reader accurately extracts data fields related to "collapse," including titles, descriptions, and keywords, thereby enhancing data discoverability. Standardizing the extracted fields into metadata documents according to the FAIR principles makes these data more easily retrievable and interoperable by machines. Additionally, the referencing information in the metadata documents enhances the reusability of the data, enabling other researchers to access and utilize the data more conveniently.

\subsection{Interoperable and Resuable}
The interoperability of data has revolutionized the landscape of large-scale data analysis, enabling machine-based processing through metadata documents. Leveraging the capabilities of the AutoFAIR, we have pioneered the development of dataset spatiotemporal mapping webpages within the realm of mountain hazard datasets. By employing a standardized metadata vocabulary, we have effectively mapped dataset descriptions onto spatiotemporal maps, thus facilitating seamless retrieval based on temporal or spatial parameters. Moreover, our extensive metadata resources empower users to filter and access desired data efficiently, whether by keywords or research institution affiliations.
\begin{figure}[h]
  \centering
  \includegraphics[width=\linewidth]{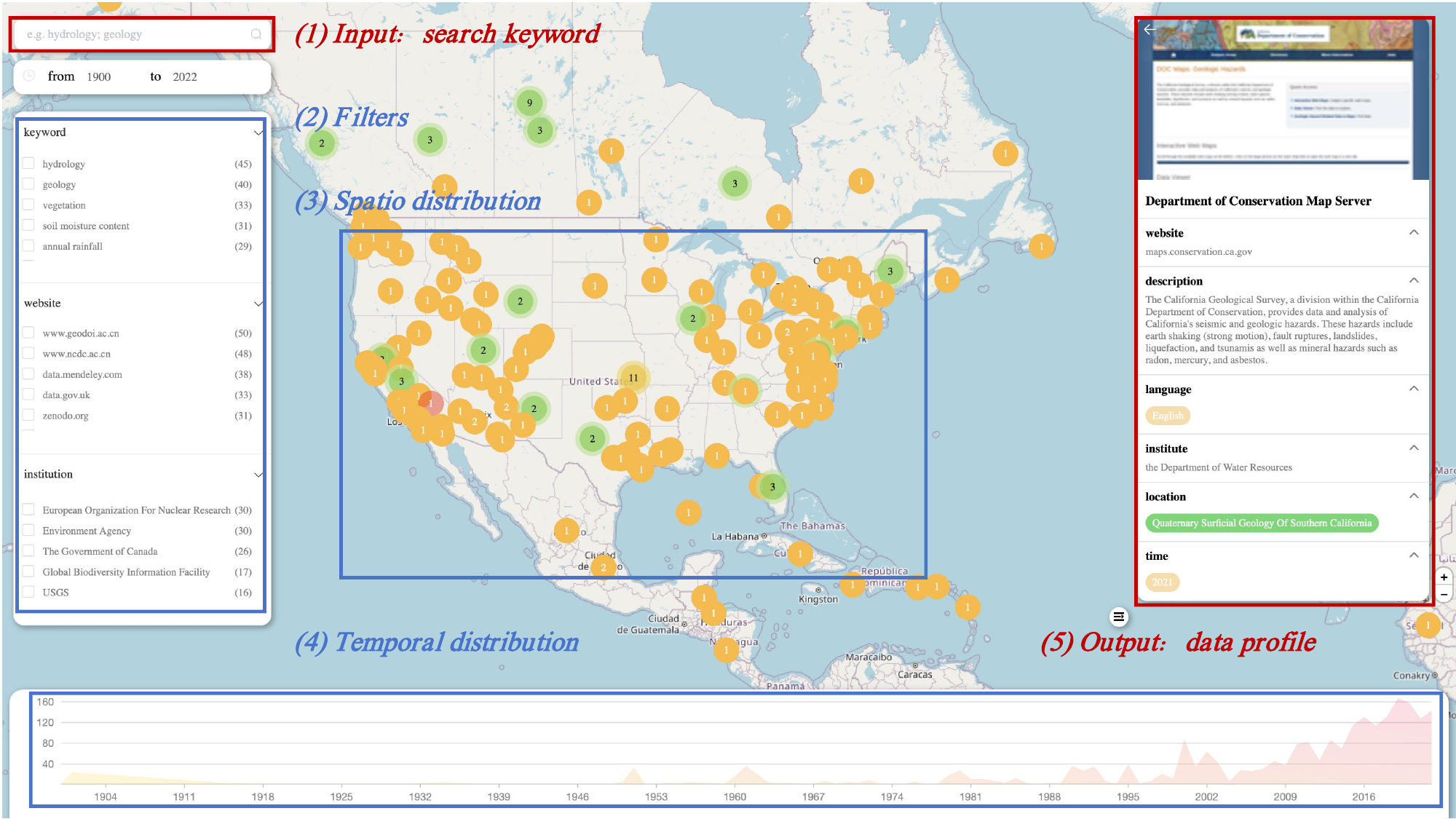}
  \caption{FAIR-compliant metadata information can provide spatial and temporal map search.}
  \label{fig:map}
\end{figure}

\begin{figure}[h]
  \centering
  \includegraphics[width=0.7\linewidth]{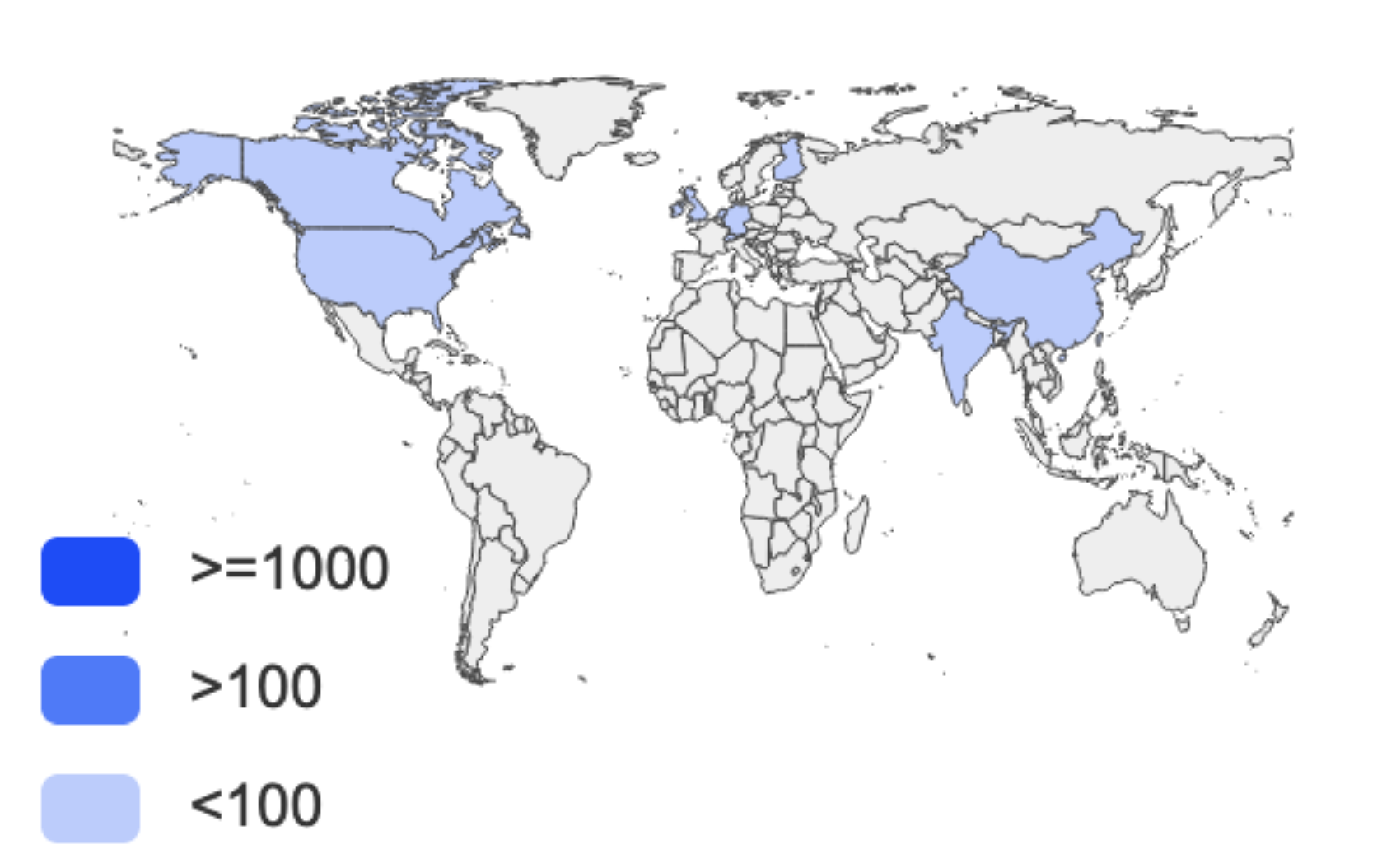}
  \caption{Spatial Distribution of Data on "Collapse". There is more open data in Europe, America and China.}
  \label{fig:spatio}
\end{figure}

\begin{figure}[h]
  \centering
  \includegraphics[width=0.7\linewidth]{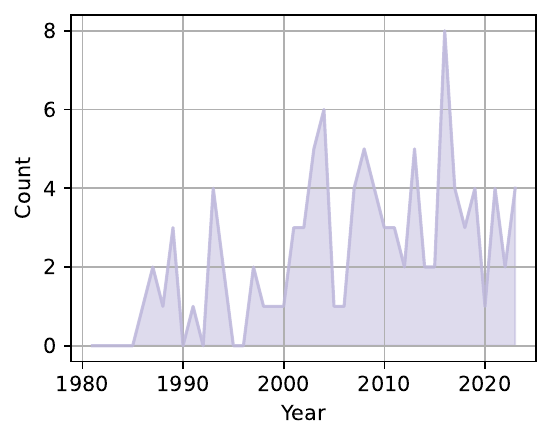}
  \caption{Temporal Distribution of Data on "Collapse". More data has emerged in recent years. }
  \label{fig:time}
\end{figure}

The spatial distribution illustrated in \autoref{fig:spatio} suggests that mountain hazard dataset contributions on "collapse" are primarily from North America and Europe, with emerging contributions from Asia, particularly China.  The temporal distribution showed in \autoref{fig:time} indicates a growing interest and accumulation of datasets on the topic of "collapse," especially in recent years, reflecting a possibly increasing focus on this research area.

\subsection{The Information Extraction Ability of Web Reader}
The effectiveness of our AutoFAIR tool depends on the information extraction capability of Web Reader. \autoref{tab:acc} demonstrates the extraction capability of Web Reader across 7124 websites, using data DOI and license fields as examples. It can be observed that Web Reader can accurately extract information from virtually all domains. 

\begin{table}[h]
  \caption{Web Reader has extracted DOIs and licenses for most data, but not all due to the lack of this information in the original data webpages.}
  \label{tab:acc}
  \begin{tabular}{ccl}
    \toprule
    Metadata & \#Dataset (Extracted) & Extraction Rate\\
    \midrule
     DOI & 7082 & 89.41\%\\
license & 7081 & 89.40\%\\
  \bottomrule
\end{tabular}
\end{table}
\begin{table}[h]
  \caption{Top Institutions Discovered by Web Reader}
  \label{tab:ins}
  \begin{tabular}{ccl}
    \toprule
    Institution & \#Dataset \\
    \midrule
     National Earth System Science Data Center& 1121\\
     National Tibetan Plateau Scientific Data Center & 1082\\
     \parbox[t]{7cm}{\centering Institute of Geographic Sciences and Natural Resources Research, Chinese Academy of Sciences} & 509\\
     National Glacier Permafrost Desert Scientific Data & 401\\
     A Big Earth Data Platform for Three Poles & 542\\
     UMS PatriNat OF CNRS MNHN & 155\\
     \parbox[t]{7cm}{\centering Institute of Tibetan Plateau Research, Chinese Academy of Sciences} & 120\\
     Plazi.org taxonomic treatments database & 116\\

  \bottomrule
\end{tabular}
\end{table}

Additionally, through the institution information extraction of Web Reader, we have successfully identified the top data research and publishing institutions in the field of mountain hazard. \autoref{tab:ins} summarizes the high-frequency institutions in the mountain hazard research domain.

\section{Conclusion}
In conclusion, AutoFAIR presents a comprehensive solution for automating data FAIRification, significantly enhancing data compliance with the FAIR principles. By integrating the Web Reader's two-stage information extraction method, which considers both DOM tree structures and textual semantics, and the FAIR Alignment module, which utilizes ontology guidance and semantic matching, AutoFAIR effectively extracts and standardizes metadata across diverse and unstructured data sources. The application of AutoFAIR in the domain of mountain hazards demonstrated notable improvements in data FAIRness, proving the system's effectiveness. Future work could extend AutoFAIR's applicability to other domains and refine its capabilities to manage complex data structures, leveraging advanced natural language processing and machine learning techniques to further optimize metadata extraction. This work contributes to the broader adoption of FAIR principles, facilitating data sharing and promoting innovation across various fields.

\begin{acks}
This work was partially supported by National Key RD Program of China (No.2022YFB3904204), National Natural Science Foundation of China (No.62272301, 623B2071, 61960206002, 62061146002, 62020106005), and Shanghai Pilot Program for Basic Research -Shanghai Jiao Tong University.
\end{acks}

\bibliographystyle{ACM-Reference-Format}
\bibliography{sample}

\end{document}